\pdfoutput=1

\documentclass[11pt]{article}

\usepackage[preprint]{acl}

\usepackage{times}
\usepackage{latexsym}

\usepackage[T1]{fontenc}

\usepackage[utf8]{inputenc}

\usepackage{microtype}

\usepackage{inconsolata}

\usepackage[most]{tcolorbox}
\usepackage{graphicx,amssymb}
\usepackage{multirow}
\usepackage{array}
\usepackage{booktabs}
\usepackage{amsfonts}
\usepackage{bbding}
\usepackage{bm}
\usepackage{color,colortbl,xcolor}
\usepackage{stfloats}
\usepackage{enumitem}
\usepackage{nccmath}
\usepackage{subfigure}
\usepackage{soul}
\usepackage{pifont}

\usepackage{tabularx}
\usepackage{listings}

\definecolor{hlgreen}{HTML}{B2D5CB}
\definecolor{hlblue}{HTML}{ADD8E6}
\definecolor{bggrey}{HTML}{5E5D65}
\definecolor{bgpink}{HTML}{CEAEB9}
\definecolor{bgblue}{HTML}{8D91AA}

\definecolor{lightred}{RGB}{222,45,38}
\definecolor{lightblue}{RGB}{49,130,189}

\title{COGENT: A Curriculum-oriented Framework for Generating Grade-appropriate Educational Content}

\author{
Zhengyuan Liu\textsuperscript{\ding{118}*},
\ Stella Xin Yin\textsuperscript{\ding{171}}\thanks{\ Equal contribution.},
\ Dion Hoe-Lian Goh\textsuperscript{\ding{171}},
\ Nancy F. Chen\textsuperscript{\ding{118}}\\
\textsuperscript{\ding{171}}Nanyang Technological University, Singapore\\
\textsuperscript{\ding{118}}Institute for Infocomm Research (I$^2$R), A*STAR, Singapore\\
\texttt{\{liu\_zhengyuan,nfychen\}@i2r.a-star.edu.sg}
}

\begin{document}
\maketitle

\begin{abstract}
While Generative AI has demonstrated strong potential and versatility in content generation, its application to educational contexts presents several challenges. Models often fail to align with curriculum standards and maintain grade-appropriate reading levels consistently. Furthermore, STEM education poses additional challenges in balancing scientific explanations with everyday language when introducing complex and abstract ideas and phenomena to younger students.
In this work, we propose COGENT, a curriculum-oriented framework for generating grade-appropriate educational content. We incorporate three curriculum components (science concepts, core ideas, and learning objectives), control readability through length, vocabulary, and sentence complexity, and adopt a ``wonder-based'' approach to increase student engagement and interest. We conduct a multi-dimensional evaluation via both LLM-as-a-judge and human expert analysis. Experimental results show that COGENT consistently produces grade-appropriate passages that are comparable or superior to human references. Our work establishes a viable approach for scaling adaptive and high-quality learning resources.
\end{abstract}

\section{Introduction}
Educational content, particularly reading materials, is considered an integral part of supporting effective learning across disciplines. Traditionally, the creation of educational materials has relied mainly on human authors. This limits scalability and adaptability when curriculum standards evolve or when diverse learning needs must be addressed at scale. Generative AI techniques, such as Large Language Models (LLMs), have demonstrated remarkable potential in various content generation \cite{achiam2023gpt,team2024gemma}. However, their application to educational contexts presents several challenges. While models can generate grammatically correct and coherent passages, they often fail to align with established curriculum standards \cite{ref_xiao,liu2024scaffolding}. Moreover, it is difficult to maintain consistent grade-appropriate reading levels, as both sentence structure and vocabulary complexity impact student comprehension and learning outcomes \cite{ref_zamanian}. STEM education poses an additional challenge of balance between science and everyday language when introducing complex and abstract concepts to younger students \cite{ref_blown,ref_gilbert}. Therefore, creating materials that effectively bridge science terminologies with real-world examples while maintaining pedagogical value requires professional knowledge and multi-dimensional efforts \cite{ref_bansiong}. 

To address these problems, here we propose a framework \textbf{C}urriculum-\textbf{O}riented \textbf{G}eneration for \textbf{E}ducational Conte\textbf{nt} (COGENT), which creates science reading materials aligned with curriculum standards and adapts to grade-specific readability requirements. This framework consists of three components: curriculum formulation, controllable content generation, and multi-dimensional evaluation. Grounded in well-established education standards such as the Next Generation Science Standards (NGSS) \cite{ngss2013next}, we build the structured guidance by linking science concepts (e.g., grades 1-5) with core ideas and their corresponding learning objectives, which creates systematic alignment with pedagogical value.
For readability control, we implement constraints on word number, vocabulary, and sentence complexity based on grade-level reading proficiency \cite{ref_flesch}. Further, inspired by inquiry-based learning \cite{ref_dewey}, we incorporate a ``wonder-based'' learning approach that transforms core scientific ideas into inquiry-driven topics to engage students with science learning and discovery. 

To comprehensively evaluate our framework and its pedagogical effectiveness, we build a multi-dimensional validation protocol and conduct quantitative analyses of the generated content across curriculum alignment, comprehensibility, and readability metrics. Based on the COGENT framework, our experiments with three representative LLMs (Gemma-2-9B, GPT-4o, Claude-3.5-Sonnet) indicate that: (1) models can follow curriculum guidance to create educational content that aligns closely with established pedagogical standards; (2) models not only maintain high comprehensibility but also demonstrate adaptability in adjusting length, vocabulary, and sentence complexity to meet grade-specific reading requirements. The findings suggest that with proper scaffolding and constraint mechanisms, LLM-based systems can serve as a complement to human expertise in educational content development, which enables access to high-quality, curriculum-aligned reading materials across diverse educational contexts. This work not only advances our understanding of how to effectively harness models for educational purposes but also establishes a foundation for future investigations into automated content generation, with broader applications for personalized learning.

\section{Related Work}
\subsection{AI-generated Content in Education}
Advancements in LLMs have accelerated the adoption of AI in educational contexts, particularly in automating traditionally time-consuming content generation tasks such as providing feedback, creating assessment materials, and generating learning recommendations \cite{ref_yan,liu2024scaffolding,liu-etal-2024-personality}. These efforts provide customized learning materials to students based on individual factors such as learning status, preferences, and goals \cite{wang2024large,liu-etal-2024-optimizing}. For example, \citet{kuo2023leveraging} demonstrated how to generate dynamic learning paths for students based on their most recent knowledge mastery assessment results. Similarly, \citet{kabir2023llm} enhances content generation by incorporating knowledge concept structures throughout the process. While these methods show promise, they mainly focus on students' own learning trajectories and knowledge structures, with little attention given to standardized curriculum frameworks. Additionally, the generated content often fails to appropriately differentiate reading levels.

To evaluate LLM-generated content, researchers combined automatic and expert analysis. For instance, \citet{ref_lee} investigated LLMs' capability in generating test questions, with both automatic evaluation and expert analysis confirming that these models can produce questions with high validity and reliability for language learning. Similarly, \citet{zelikman-etal-2023-generating} developed a reading comprehension exercise generation system for middle school English learners, demonstrating that AI-generated materials can not only meet students' learning needs but, in some cases, surpass the quality of human-written materials. In computer science education, \citet{ref_lee2} examined the effectiveness of AI-generated content in explaining programming concepts, further validating the potential of LLMs in educational content creation.

While current evaluation of AI-generated content focuses mainly on language and facts \cite{ref_xiao}, real-world educational assessment requires broader criteria including curriculum alignment, pedagogical scaffolding, and grade-level appropriateness \cite{ref_bansiong,ref_berndt}. This lack of comprehensive evaluation standards hinders educators' interest and trust in implementing AI-generated resources.

\subsection{Evaluation Metrics of Education Materials}
The evaluation of educational content includes three aspects: readability, comprehensibility, and curriculum alignment. These factors collectively determine whether learning materials are ``appropriate to the student's age and level of knowledge'' and ``prepared in line with the curricula.''

\textit{Comprehensibility} and \textit{Readability} serve as fundamental metrics in analyzing educational texts \cite{ref_zamanian}. Readability is a textual characteristic that measures how easily text can be read and understood \cite{ref_klare}, while comprehensibility reflects how effectively readers can construct meaning from the text \cite{ref_sadoski,ref_beck}. As \citet{ref_lakoff} emphasizes, ``understanding is only possible through the negotiation of meaning.'' When these aspects are misaligned, students may experience frustration or disengagement \cite{ref_bansiong}.

\textit{Curriculum alignment} aims to ensure it meets educational standards while remaining appropriate for learners' grade levels \cite{ref_anderson}. This evaluation ensures that educational materials are not only readable and comprehensible but also serve their intended pedagogical purposes within the educational framework \cite{ref_squires,ref_wijngaards}.

\subsection{Value of ``Wonder'' in Science Education}

\textit{``The most beautiful thing we can experience is the mysterious. It is the source of all true art and science.'' (Einstein, 1931)}

Inquiry-based learning is rooted in the work of \citet{ref_dewey}, who underlines that education begins with the curiosity of the learner. Inquiry is understood in two ways: (1) ``inquiry as means'' (inquiry in science) refers to using inquiry as an instructional approach to help students develop their understanding of science content; (2) ``inquiry as ends'' (inquiry about science) refers to inquiry as a learning outcome \cite{ref_nrc, ref_abdelkhalick}.
However, when students inquire about scientific knowledge, they often experience a gap between their intuitive comprehension and their ability to express understanding \cite{ref_blown}. They frequently struggle to express their observations and questions using scientific language. This disconnect highlights the need for level-appropriate educational content that can bridge the gap between students' intuitive understanding and formal scientific language. Given this challenge, it is recommended to introduce scientific concepts through ``wonder why'' questions that trigger children's natural curiosity while reducing the barriers of science terminologies \cite{ref_chin,ref_gilbert}. Moreover, wonder-based explanatory texts are effective for reading comprehension, science learning, and conceptual change \cite{ref_lindholm,ref_jirout}.

\begin{figure*}[t!]
\centering
\includegraphics[width=1.0\textwidth]{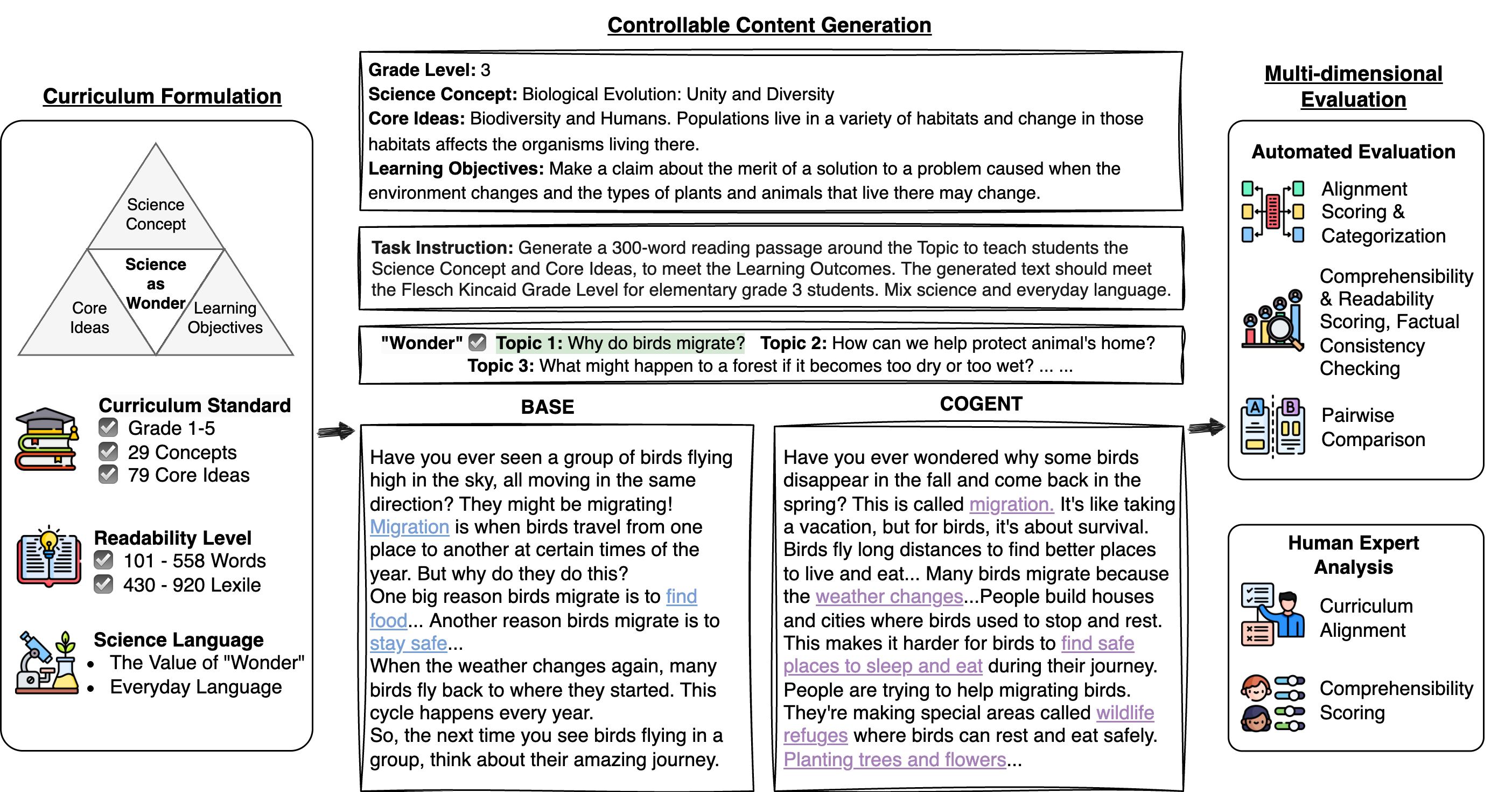}
\caption{Overview of the framework of curriculum-oriented generation for educational content (COGENT).}
\label{Fig.framework}
\end{figure*}

\section{Curriculum-Oriented Generation for Educational Content}
The framework is designed to transform abstract curriculum components into engaging, wonder-based reading materials that improve students' understanding while adhering to grade-specific readability requirements. It consists of three parts: curriculum formulation, controllable content generation, and multi-aspect evaluation (see Figure \ref{Fig.framework}).

\subsection{COGENT-based Generation}
\begin{figure*}[t!]
\centering
\includegraphics[width=1.0\textwidth]{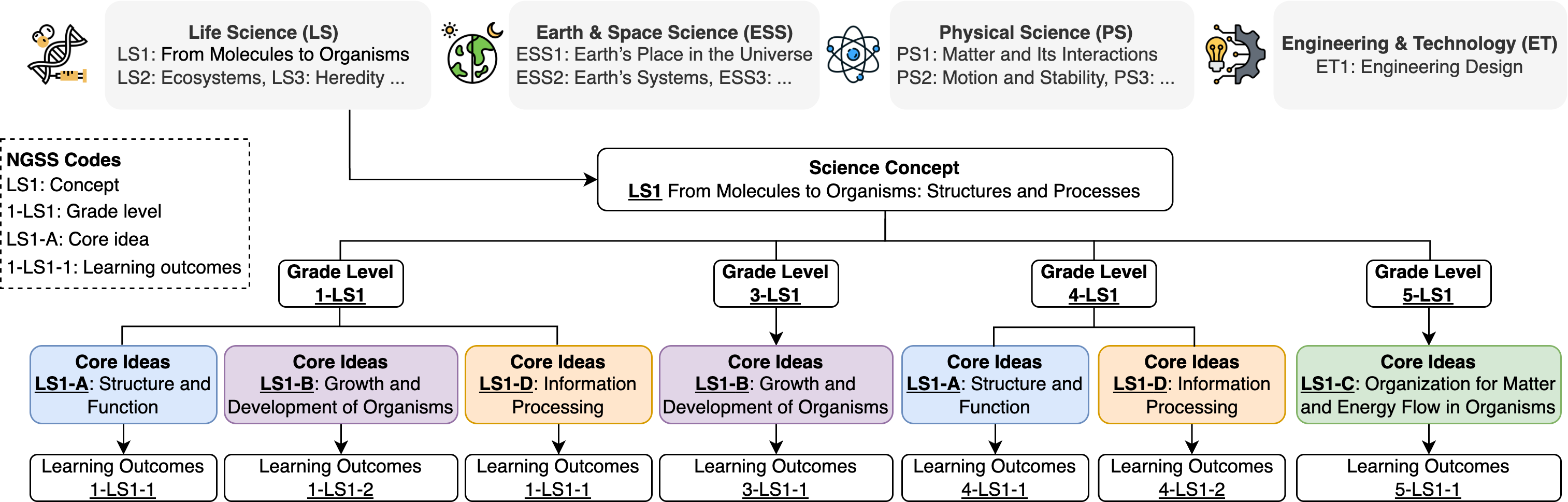}
\caption{Our curriculum decomposition example grounded in the Next Generation Science Standards (NGSS), which consist of four domains. It has a hierarchical structure where Science Concepts (e.g., LS1) branch into Core Ideas (e.g., LS1-A: Structure and Function), which then connect to specific Learning Outcomes for each grade level (e.g., 1-LS1-1). The same core idea may appear across multiple levels with increasing complexity. For example, LS1-A (Structure and Function) progresses from grade 1 to grade 4.}
\label{Fig.NGSS}
\vspace{-0.3cm}
\end{figure*}

To simulate human teachers and editors \cite{bybee2014ngss}, we incorporate structured curriculum information to guide LLM-based educational content generation, ensuring pedagogical alignment, development progress, and topic coverage.
Here, we ground our approach in the Next Generation Science Standards (NGSS), a well-established K-12 science education framework \cite{ngss2013next}.\footnote{While we demonstrate our framework using NGSS as a representative example in this paper, the hierarchical decomposition underlying COGENT can be adapted to other national education frameworks and subjects, such as the National Curriculum in England \cite{dfe2014national_collection} or Singapore's Ministry of Education curriculum standards \cite{moe2023primary}.}
We decompose the curriculum into three hierarchical elements: science concepts, core ideas, and learning objectives. As shown in Figure \ref{Fig.NGSS}, science concepts can be mapped to core ideas, and each core idea is related to learning outcomes, creating a comprehensive curriculum coverage matrix. 
More specifically, for elementary school students (grades 1-5, ages 6-11), 29 science concepts (e.g., \textit{``Matter and Its Interactions''}) are broken down into 79 core ideas (e.g., ``\textit{Structure and Properties of Matter. Matter can be described and classified by its observable properties.}''), then further mapped to specific learning outcomes that detail what students should master at each grade level (e.g., ``\textit{To describe and classify different kinds of materials by their observable properties (Grade 2).}'').

\begin{table}[t!]
\centering
\resizebox{1.0\linewidth}{!}
{
\begin{tabular}{lccc}
\toprule
\textbf{Level} & \textbf{Avg. words} & \textbf{Avg. lexile} & \textbf{Avg. unique words} \\
\midrule
Grade 1 (Ages 6-7) & 101 & 430 & 57.9 \\
Grade 2 (Ages 7-8) & 200 & 545 & 87.7 \\
Grade 3 (Ages 8-9) & 319 & 605 & 132.8\\
Grade 4 (Ages 9-10) & 468 & 770 & 183.2\\
Grade 5 (Ages 10-11) & 558 & 920 & 219.5\\
\bottomrule
\end{tabular}
}
\caption{Linguistic features of human-written science reading passages at elementary grade levels.}
\label{table-grade-diff}
\vspace{-0.2cm}
\end{table}

Importantly, concepts and core ideas can appear across multiple grade levels, requiring different depths of explanation and language complexity (see Figure \ref{Fig.NGSS}). As shown in Table \ref{table-grade-diff}, human-written science reading passages show clear patterns across grade levels: the average number of words, reading difficulty scores (lexile) \cite{white2001assessing}, and lexical diversity all increase steadily as students progress from grade 1 to grade 5. We thus indicate the word number and target readability level \cite{ref_klare,ref_flesch}\footnote{In our experiments, based on human-written passages, we set the word count to be the grade level multiplied by 100. Flesch Kincaid Grade Level is used for readability control.}  along with the curriculum input to ensure generated content matches students' reading abilities at each grade.

Moreover, to enhance students' interest and engagement, we consider ``Science as Wonder'' and ``everyday language'' as a bridge to connect scientific concepts and their daily experiences. 
Given the decomposed curriculum items, each core idea can be used to generate multiple exploration questions. For example, the core idea about environmental adaptation can be linked to wonder topics such as \textit{``Why do birds migrate?''} or \textit{``How can we help protect animals' home?''} This approach maintains curriculum alignment while fostering student curiosity through diverse and interesting content. When explaining bird migration, the generated passage begins with an interesting observation (``\textit{Some birds disappear in the fall and come back in the spring.}''), followed by clear explanations of stories and scientific concepts, and concludes with broader implications for environmental understanding.

\subsection{Multi-dimensional Evaluation}
While LLM-generated content can be modulated along desired dimensions to meet specific requirements, it may not perform consistently and precisely \cite{ref_saha,li-etal-2025-dna}. We thus propose a multi-dimensional evaluation to validate pedagogical effectiveness and generation quality.

First, we evaluate \textbf{Curriculum Alignment} through scoring and categorization schemes. The scoring evaluates how well the content adheres to the specified curriculum item, and the categorization examines whether the passage delivers exact core ideas and outcomes at each grade level. Evaluation examples are shown in Table \ref{table-inst-alignment}.\\
\noindent \textit{\textbf{Curriculum Alignment Scoring}}: We rate the passage compliance with the standards using a 5-point scale (1 = does not align at all, 5 = fully aligned). Given a sample set, we calculate the average score to determine its overall curriculum alignment.\\
\noindent \textit{\textbf{Curriculum Item Categorization}}: Since science concepts appear in multiple grade levels, we first group passages by concept (e.g., \textit{``From Molecules to Organisms: Structures and Processes''}), and classify them into the corresponding curriculum item: a tuple of $\{$\textit{concept, core idea, learning outcome}$\}$. For example, as shown in Figure \ref{Fig.NGSS} and Table \ref{table-inst-alignment}, the input passage will be classified into one of the seven types (e.g., ``\textit{Type A (core idea): Structure and Function. All organisms have external parts}'', ``\textit{Type G: Organization for Matter and Energy Flow in Organisms}'').

We then evaluate the \textbf{Comprehensibility} from four aspects following previous work \cite{ref_celikyilmaz}. This is to show how effectively readers can construct meaning from the text. Each dimension is in a 5-point Likert scoring:
\textit{Readability} (How easily the text can be read and understood), \textit{Correctness} (The accuracy of factual content about the topic), \textit{Coherence} (The consistency between the content and the topic), and \textit{Engagement} (To what extent the ``wonder-based'' topic and passage capture and maintain readers' interest). Examples can be found in Table \ref{table-inst-comprehensibility}.

Moreover, we use four common statistical methods to assess \textbf{Text Readability} based on linguistic features: \textit{Flesch Reading Ease/Flesch Kincaid Grade Level} \cite{ref_flesch} evaluates readability using sentence length and syllable count, with scores from 0-100 (higher meaning easier to read) or converted to grade levels. \textit{Gunning Fog Index }\cite{ref_gunning} measures complexity through sentence length and percentage of complex words, indicating education years needed for comprehension. \textit{Automated Readability Index} \cite{ref_smith} and \textit{Coleman Liau Index} \cite{ref_liau} differ from other formulas by using character count instead of syllable count, along with average word and sentence length (see examples in Table \ref{table-inst-readability}).

\section{Experimental Setting}
We conducted extensive experiments on science reading passage generation to examine both the effectiveness and pedagogical value of COGENT. Since this task requires structured instruction following and coherent language generation, we applied and tested three representative LLMs: Gemma-2-9B-IT \cite{team2024gemma}, GPT-4o\footnote{https://platform.openai.com/docs/models/gpt-4o} (version 20240806), and Claude-3.5-Sonnet\footnote{https://docs.anthropic.com/en/docs/about-claude/models/all-models} (version 20241022). We use the default generation parameters (e.g., temperature, top-p) in their model configurations. The example instructions for wonder question generation, and BASE and COGENT passage generation are shown in Table \ref{table-prompt-example}.


\subsection{Comparison through Grouped Generation and Human-written Passages}
First, we collect and assess grouped passages generated from the same curriculum inputs to evaluate COGENT's capability in generating diverse yet consistent content. Given each $\{$\textit{concept, core idea, learning outcomes}$\}$ tuple, we randomly generated three ``wonder'' topics, then created corresponding reading passages for each topic.

Moreover, we collect 50 human-written passages and build an evaluation set for extensive comparison. These passages were selected from verified educational resources and textbooks, covering various science concepts across elementary school grades 1-5. Each sample was annotated with corresponding curriculum standards and readability metrics, which provide a high-quality reference.

\begin{table}[t!]
\centering
\small
\resizebox{1.0\linewidth}{!}
{
\begin{tabular}{ccccc}
\toprule
\textbf{Grade} & \textbf{Type} & \textbf{Gemma-2} & \textbf{GPT-4o} & \textbf{Claude-3.5} \\
\midrule
1 & BASE & 91.13 & 110.30 & 98.10 \\
1 & COGENT & 82.03 & 113.30 & 99.17 \\
2 & BASE & 151.13 & 206.13 & 204.54 \\
2 & COGENT & 119.85 & 193.13 & 199.69 \\
3 & BASE & 250.63 & 336.61 & 290.33 \\
3 & COGENT & 215.44 & 311.09 & 292.67 \\
4 & BASE & 350.50 & 468.77 & 404.86 \\
4 & COGENT & 365.53 & 418.23 & 395.09 \\
5 & BASE & 418.23 & 590.21 & 518.63 \\
5 & COGENT & 387.21 & 556.19 & 492.00 \\
\bottomrule
\end{tabular}
}
\caption{Statistics of the generation length.}
\label{tab:word-num-compare}
\vspace{-0.2cm}
\end{table}

\subsection{Evaluation Methods and Process}
For automated evaluation, we leverage LLM-as-a-judge for automated scoring on the \textbf{Curriculum Alignment} and \textbf{Comprehensibility} scoring \cite{ref_saha}. In our preliminary testing, Claude-3.5-Sonnet performs well as a consistent and accurate evaluator. To assess the grouped generation, we reported the average scores of three passages per topic to reduce intrinsic bias from the LLM-based annotator. We use an off-the-shelf tool to calculate \textbf{Text Readability} scores.\footnote{https://github.com/textstat/textstat}
Moreover, for curriculum item categorization, we group the 79 core ideas based on their science concepts and classify samples within each group. The accuracy is an indicator to measure the distinctness of grade-specific explanation depth and learning objectives.

\begin{table*}[t!]
\centering
\small
\resizebox{1.0\linewidth}{!}
{
\begin{tabular}{lp{7.5cm}ccc}
\toprule
\textbf{Metric} & \textbf{Description} & \textbf{BASE} & \textbf{COGENT} & \textbf{\textit{p}-value} \\
\midrule
Curriculum Alignment & How well content aligns with curriculum standards & 4.08 & 4.62 & \textcolor{red}{.021*} \\
Comprehensibility & How effectively readers can construct meaning from the text (readability, correctness, coherence, and engagement) & 4.76 & 4.81 & .083 \\
\bottomrule
\end{tabular}
}
\caption{Statistical comparison of curriculum alignment and comprehensibility metrics: BASE vs COGENT. \textit{p}-value is calculated through pairwise Mann-Whitney U tests with Bonferroni correction (** \textit{p} <.01, * \textit{p} <.05).}
\label{tab:comparison_2}
\vspace{-0.1cm}
\end{table*}

\begin{figure*}[t!]
\centering
\includegraphics[width=1.0\textwidth]{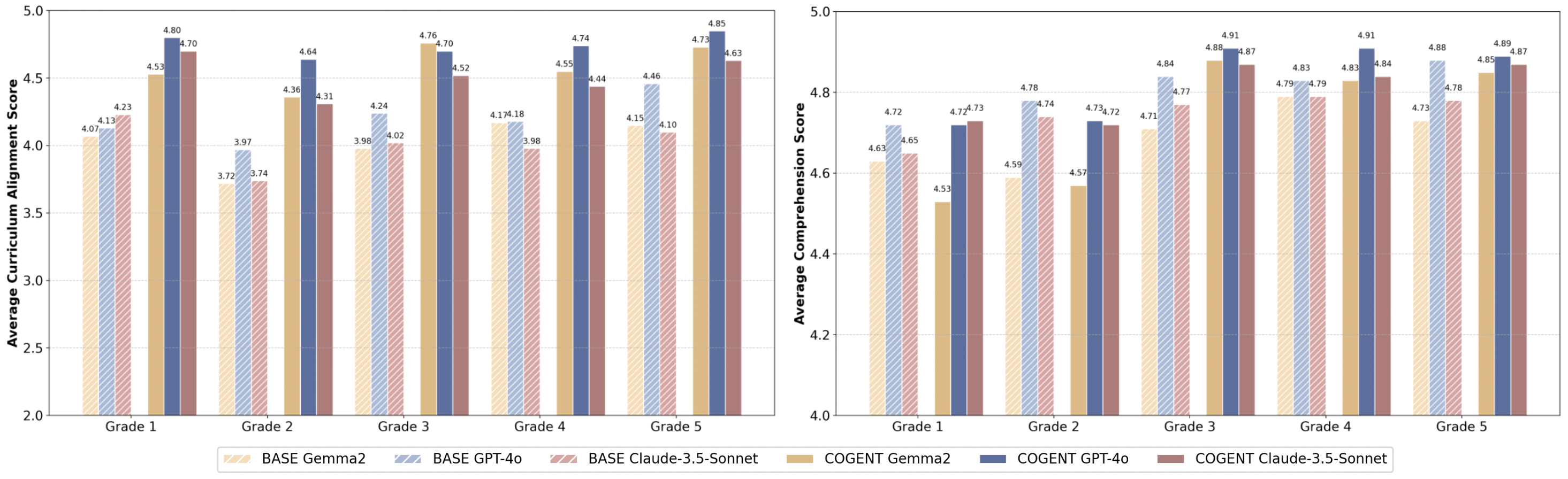}
\caption{Curriculum alignment scores (left) and comprehensibility scores (right) of Gemma-2-9B, GPT-4o, and Claude-3.5-Sonnet generated passages using BASE and COGENT framework.}
\label{Fig.Curriculum_all}
\vspace{-0.1cm}
\end{figure*}

\begin{figure*}[t!]
\centering
\includegraphics[width=1.0\textwidth]{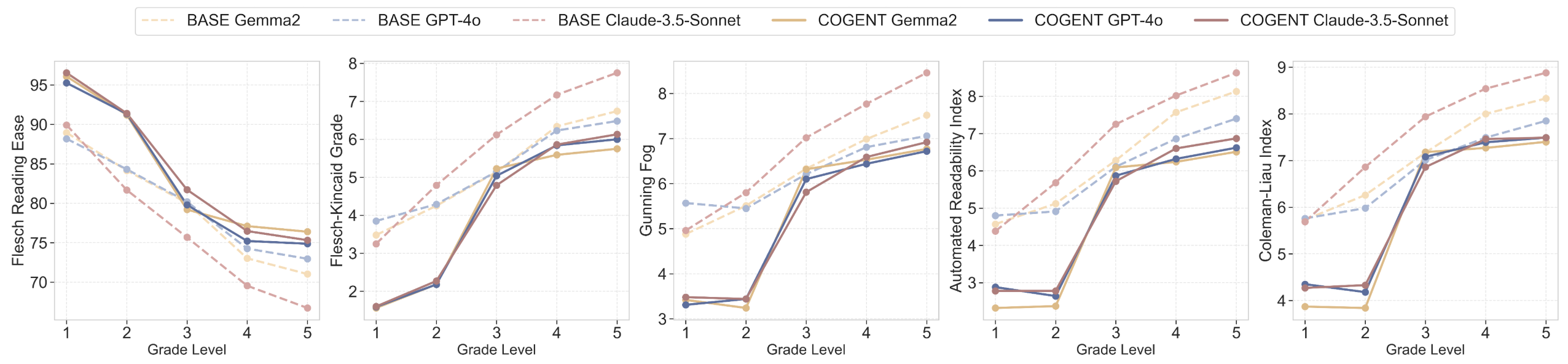}
\caption{Results on four readability metrics of LLM-generated passages using BASE and COGENT framework.}
\label{Fig.Readability_all_metric}
\vspace{-0.1cm}
\end{figure*}

For expert analysis, we recruited six elementary science teachers who have more than 10 years' teaching experience to conduct expert analysis. Teachers evaluated passages from grades 1-5, with each grade having three passages: human-written, BASE-generated, and COGENT-generated. The human evaluation consists of two surveys: \textbf{Curriculum Alignment} survey requires teachers to indicate their agreement on whether the passages aligned with corresponding grade-level science concepts and core ideas, and \textbf{Comprehensibility} survey requires them to rate each passage on four dimensions (readability, correctness, coherence, and engagement). Both surveys used the same items as the LLM-as-a-judge evaluation.

\section{Experimental Results and Discussions}
\subsection{Results on Grouped Generation}
\label{ssec:grouped_generation}
In our experiments, we generated passages (three samples per curriculum item) with Gemma-2-9B, GPT-4o, and Claude-3.5-sonnet; the total number is 711. For the \textbf{Curriculum Alignment} scoring, we conducted Mann-Whitney U tests, and the results reveal significant improvements between BASE and COGENT frameworks (see Table \ref{tab:comparison_2}). More specifically, COGENT (\textit{Mean} = 4.62) achieves significantly higher alignment scores compared to BASE (\textit{Mean} = 4.08) (\textit{p} < .05), indicating that COGENT effectively incorporates curriculum information into generated passages. As shown in Figure \ref{Fig.Curriculum_all} (left), models with COGENT demonstrate higher scores across all grade levels. While Gemma-2-9B is in a smaller parameter size, it can provide reasonable outputs following the curriculum condition, and GPT-4o performs slightly better.

Meanwhile, results of \textbf{Curriculum Item Categorization} also demonstrate COGENT's effectiveness on pedagogical alignment. For each model, we calculated and averaged the classification accuracy on 237 samples. GPT-4o achieves 0.785 with COGENT guidance, a 20\% improvement compared to 0.654 of the BASE. Similarly, Claude-3.5 improves from 0.616 to 0.726 (17.8\% relative gain) and Gemma-2 improves from 0.633 to 0.747. These improvements suggest that LLMs can follow the curriculum guidance to effectively reflect grade-specific content and objectives.

\begin{table*}[t!]
\centering
\small
\resizebox{1.0\linewidth}{!}
{
\begin{tabular}{p{3.1cm}w{c}{1.4cm}w{c}{1.4cm}w{c}{1.4cm}w{c}{2.3cm}w{c}{2.3cm}w{c}{2.5cm}}
\toprule
\textbf{Metric} & \textbf{BASE} & \textbf{COGENT} & \textbf{Human} & \textbf{BASE vs COGENT} & \textbf{BASE vs  Human} & \textbf{COGENT vs  Human} \\
\midrule
Curriculum Alignment & 3.23 & \textbf{4.15} & 3.49 & \textcolor{red}{.008**} & .067 & \textcolor{red}{.029*} \\
Comprehensibility & 4.47 & \textbf{4.58} & 4.16 & .053 &\textcolor{red}{.022*} & \textcolor{red}{.014*} \\
\bottomrule
\end{tabular}
}
\caption{Statistical comparison of curriculum alignment and comprehensibility: BASE vs COGENT vs Human \textit{p}-value is calculated through pairwise Mann-Whitney U tests with Bonferroni correction (** \textit{p} <.01, * \textit{p} <.05).}
\label{tab:comparison_all3}
\vspace{-0.1cm}
\end{table*}

\begin{figure*}[t!]
\centering
\includegraphics[width=1.0\textwidth]{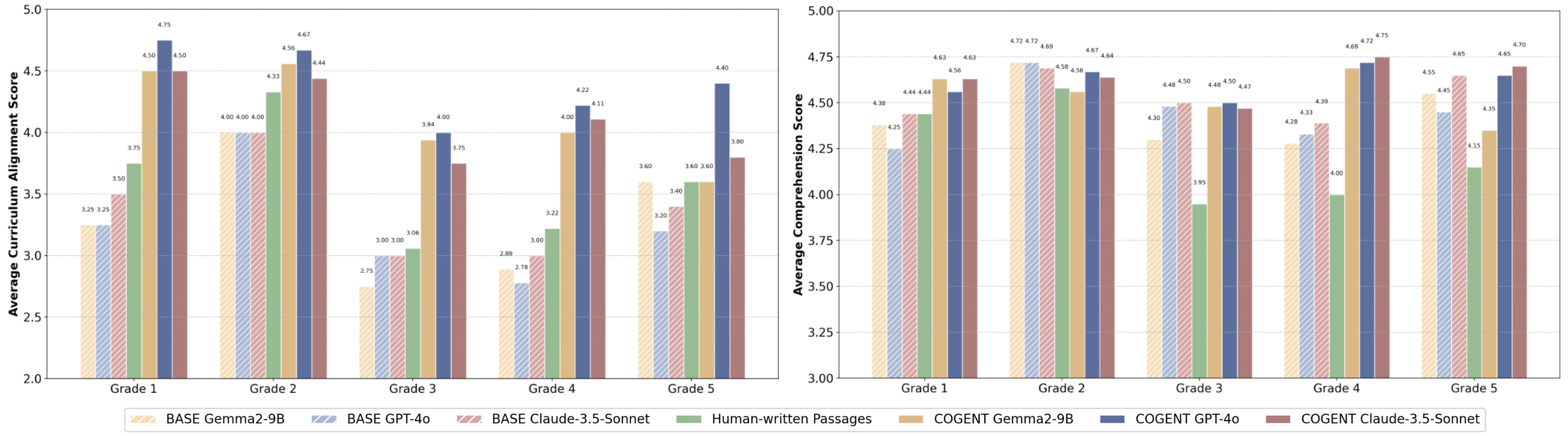}
\caption{Results on curriculum alignment and comprehensibility of Human, BASE, and COGENT.}
\label{Fig.Curriculum_human}
\vspace{-0.1cm}
\end{figure*}

\begin{figure*}[t!]
\centering
\includegraphics[width=1.0\textwidth]{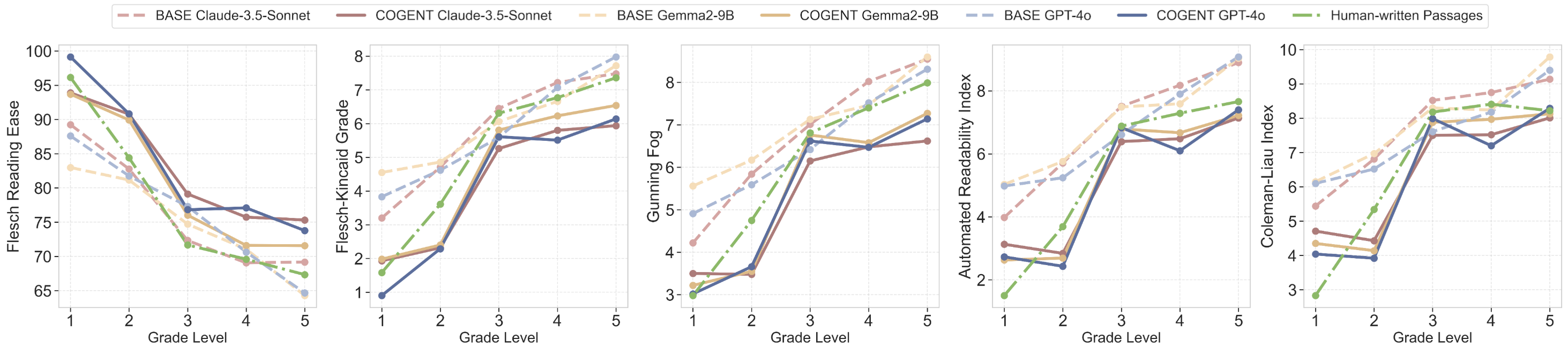}
\caption{Results on readability metrics of human-written passages, BASE, and COGENT framework.}
\label{Fig.Readability_human}
\vspace{-0.3cm}
\end{figure*}

Regarding \textbf{Comprehensibility}, models with BASE and COGENT perform well and comparable (4.76 vs 4.81) (\textit{p} = .083), as shown in Table \ref{tab:comparison_2}; they do not have significant variance across grade levels, as shown in Figure \ref{Fig.Curriculum_all} (right). This demonstrates that adding curriculum targets in the science reading passages does not affect the ease of comprehension. Moreover, we observed that tested LLMs perform well (<6\% averaged error rate) regarding \textbf{Factual Correctness} on the elementary school content writing \cite{HughesBae2023}.

\begin{table}[t!]
\centering
\small
\begin{tabular}{w{c}{0.6cm}w{c}{1.5cm}w{c}{1.9cm}w{c}{1.9cm}}
\toprule
{\textbf{Grade}} & {\textbf{Human}} & {\textbf{BASE}} & {\textbf{COGENT}} \\
\midrule
1 & 57.9 & \cellcolor{lightred!20}66.5 \textsuperscript{(+14.8\%)} & \cellcolor{lightred!20}66.5 \textsuperscript{(+14.8\%)} \\
2 & 87.7 & \cellcolor{lightred!40}110.6 \textsuperscript{(+26.1\%)} & \cellcolor{lightred!30}100.7 \textsuperscript{(+14.9\%)} \\
3 & 132.8 & \cellcolor{lightred!20}153.2 \textsuperscript{(+15.3\%)} & \cellcolor{lightred!10}137.1 \textsuperscript{(+3.2\%)} \\
4 & 183.2 & \cellcolor{lightred!10}196.1 \textsuperscript{(+7.0\%)} & \cellcolor{lightblue!10}174.0 \textsuperscript{(-5.0\%)} \\
5 & 219.5 & \cellcolor{lightred!10}230.5 \textsuperscript{(+5.0\%)} & \cellcolor{lightblue!10}209.0 \textsuperscript{(-4.8\%)} \\
\bottomrule
\end{tabular}
\caption{Comparison of unique words. Red and blue indicate the intensity of higher and lower scores compared with human-written passages, respectively.}
\label{tab:heatmap}
\vspace{-0.5cm}
\end{table}

We observed that LLMs are well-conditioned on the word count (see Table \ref{tab:word-num-compare}) at all grade levels. This ability to control length is important for creating grade-appropriate passages, as it is one of the factors that affect readability. However, on statistical \textbf{Text Readability} metrics, the two approaches perform differently. Results in Figure \ref{Fig.Readability_all_metric} show that COGENT adheres more closely to elementary reading levels, especially in lower grades (e.g., 1-2), where the BASE approach exceeds the intended level by around 2.5 grades.
The above results highlight the distinction between readability (e.g., word count and sentence complexity) and actual comprehension ease, which depends on factors like coherence, engagement, and contextual clarity \cite{ref_bansiong}. Without curriculum information, LLMs are prone to produce content beyond the indicated grade level, and grade-appropriate generation should meet both requirements.

\begin{figure*}[t!]
\centering
\includegraphics[width=1.0\textwidth]{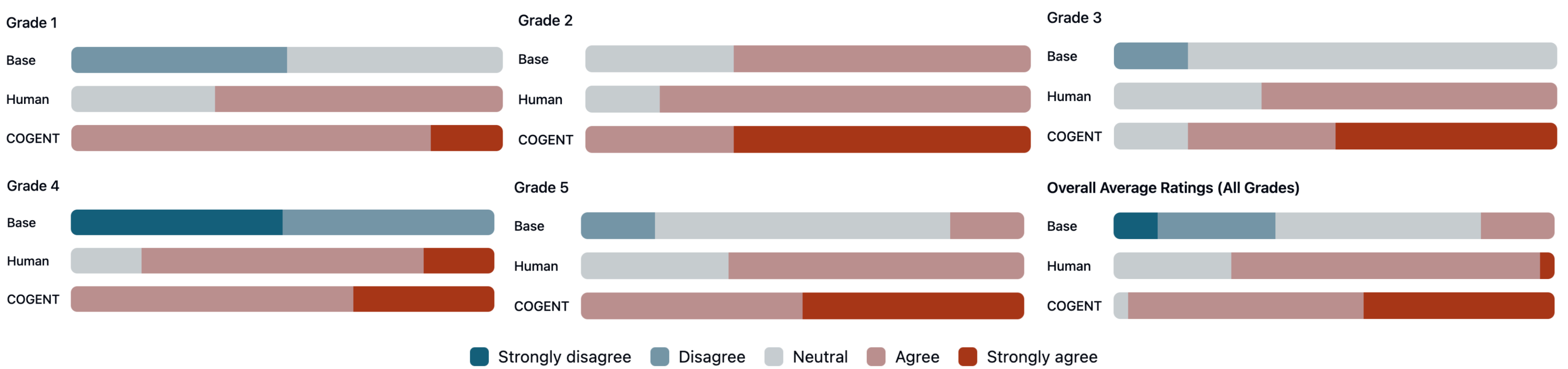}
\caption{Expert analysis: curriculum alignment comparison of Human, BASE, and COGENT.}
\label{Fig.Human}
\vspace{-0.2cm}
\end{figure*}

\begin{figure*}[t!]
\centering
\includegraphics[width=0.97\textwidth]{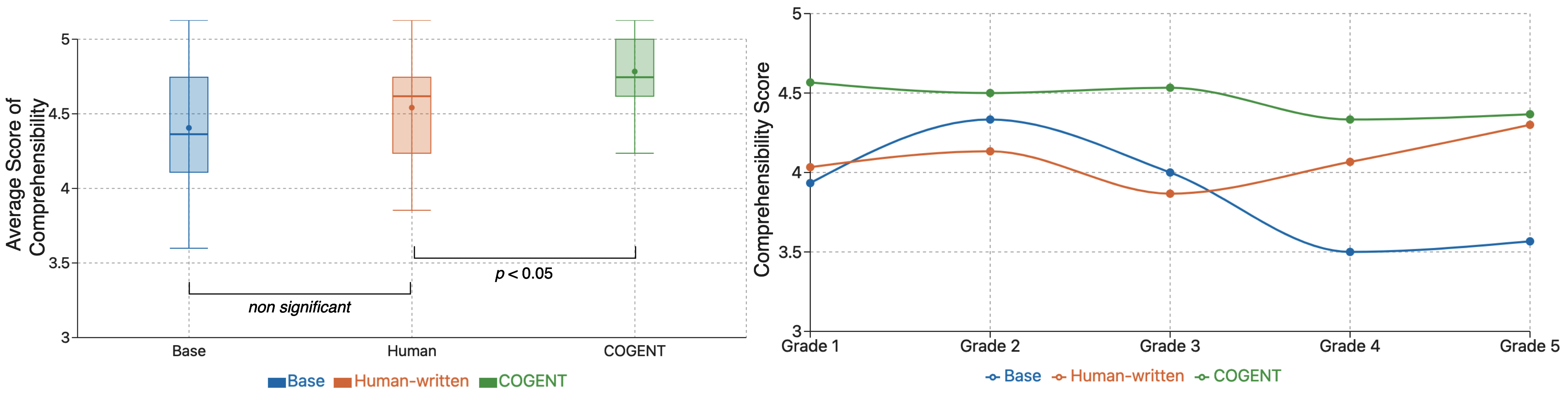}
\caption{Expert analysis: comprehensibility score comparison among Human, BASE, and COGENT.}
\label{Fig.Human2}
\vspace{-0.3cm}
\end{figure*}

\subsection{Comparison to Human-written Passages}
We used the same wonder topics and word numbers as the 50 human-written passages for a parallel comparison.
Table \ref{tab:comparison_all3} shows Mann-Whitney U test results among BASE, COGENT, and Human. We observe substantial improvement in \textbf{Curriculum Alignment}, and comparable scores in \textbf{Comprehensibility}.
COGENT demonstrates much higher alignment scores (\textit{Mean} = 4.15) than both BASE (\textit{Mean} = 3.23) and Human (\textit{Mean} = 3.49) (\textit{p} < .05). Similar to grouped generation (Section \ref{ssec:grouped_generation}), COGENT achieves better alignment scores at all grades. This indicates that COGENT-guided passages align better with curriculum standards. Among the three LLMs, GPT-4o results in slightly higher scores (see Figure \ref{Fig.Curriculum_human}). Surprisingly, Human, BASE, and COGENT all receive lower alignment ratings in grades 3-5. This occurs because the wonder topics extracted from the human references are not well-matched in these higher grades.

Second, \textbf{Comprehensibility} evaluation results show that both BASE (\textit{Mean} = 4.47) and COGENT (\textit{Mean} = 4.58) outperform Human (\textit{Mean} = 4.16) (\textit{p} < .05), while the difference between COGENT and BASE is not statistically significant. Interestingly, all three approaches maintain relatively high comprehensibility scores, while human-written passages show a notable decline from grade 3. There is a similar trend in readability evaluation results.

Third, \textbf{Text Readability} assessment results demonstrate that COGENT's performance more closely correlates with human references, although the latter slightly exceeds target grade levels.
As shown in Figure \ref{Fig.Readability_human}, on the linguistic metrics, COGENT produces passages closer to the intended grade level, while BASE generates passages largely above intended grade levels. For example, when targeting grade 1 content, BASE produces text at grade 3-4 reading level, which creates potential comprehension barriers for early readers. Interestingly, we notice a sharp increase in difficulty level at grade 3, which represents the significant transition in science education at this level. In grade 2, science learning focuses on concrete concepts through basic observation, classification, and simple investigations of the natural world, while starting from grade 3, teachers introduce more complex scientific concepts requiring deeper analysis and abstract thinking.

We also calculate the unique word numbers of each passage created by Human, BASE, and COGENT. Both BASE and COGENT show higher vocabulary diversity than human writing in early grades, with BASE producing up to 26.1\% more unique words at grade 2. This gap narrows in higher grades, where BASE still generates more unique words ($\mathrm{+}$5-7\%), while COGENT shifts to slightly lower lexical diversity ($\mathrm{-}$5\%) than human writing. The trend suggests that COGENT vocabulary usage becomes more aligned with human patterns as grade levels increase.

\subsection{Expert Analysis}
We conducted expert analysis by comparing automated approaches (w/ GPT-4o) and human reference (15 reading passages). As shown in Figure \ref{Fig.Human}, \textbf{Curriculum alignment} results align with our previous evaluation findings. COGENT achieves consistently higher alignment scores. In contrast, human-written passages maintain moderate alignment across all grades, while the BASE shows declining alignment scores in higher grades. At each grade level, COGENT maintains the highest proportion of positive ratings. Human-generated content generally receives favorable evaluations. BASE shows the most inconsistent performance, with a particularly lower rating at grade 4.

Regarding \textbf{Comprehensibility} (see Figure \ref{Fig.Human2}), experts assigned the highest ratings to COGENT-generated passages, with significant difference compared to human-written passages (\textit{p} < .05). Interestingly, BASE-generated passages and human-written passages exhibit similar comprehensibility levels in lower grades; however, their performance diverges significantly from grade 3. This divergence suggests that as grade levels increase and science concepts become more complex and abstract, the BASE framework fails to maintain appropriate readability, coherence, and engagement levels. In contrast, our framework maintains consistent comprehensibility scores at all grade levels. This highlights that based on our COGENT framework, LLM-generated reading materials achieve comparable or superior quality compared with human-authored passages, and they can be a reasonable supplement to meet both curriculum alignment and readability requirements.

\section{Conclusion}
We presented COGENT, a curriculum-oriented framework for generating grade-appropriate educational content by incorporating structured curriculum components (e.g., concepts, core ideas, and learning objectives) alongside controlled readability parameters and the ``wonder-based'' inquiry approach. Extensive experiments with three LLMs and expert evaluations demonstrate that COGENT significantly improves curriculum alignment, maintains high comprehensibility while controlling text readability to match grade levels, and generates passages comparable or superior to human-written passages. These findings establish that properly guided LLMs can serve as effective tools for scaling adaptive learning resources, with implications for educational equity and accessibility. Since COGENT is a general framework, future work could explore fine-grained personalization, interdisciplinary applications, and long-term learning outcomes to further enhance automated educational content generation.

\section*{Limitations}
While this study advances the practical application of LLMs, it has some potential limitations that warrant future study. First, our framework focused on elementary education (grades 1-5); future work could extend it to middle and high school curricula and adapt the evaluation metrics for more complex science concepts. Second, we did not include elementary students in our sample analysis due to several considerations: their limited subject knowledge and lack of understanding of curriculum standards would affect their ability to evaluate quality. Additionally, in readability assessments, younger students tend to focus on surface-level features (like pictures and length) rather than the accuracy of scientific content, clarity of explanations, or scaffolding of complex ideas. These could potentially introduce bias in the assessment results.

\section*{Ethics and Impact Statement}
We acknowledge that all of the co-authors of this work are aware of the provided ACL Code of Ethics and honor the code of conduct. In our experiments, models are applied under proper license. All data used in this work are only for academic research purposes and should not be used outside of academic research contexts. Our proposed methodology, in general, does not create a direct societal consequence and is intended to be used to improve accessibility and educational value.

\section*{Acknowledgments}
This research is supported by the AI4EDU Programme in the Institute for Infocomm Research (I$^2$R), Agency for Science, Technology and Research (A*STAR), Singapore. We thank the anonymous reviewers for their precious feedback to help improve and extend this piece of work.

\bibliography{custom}

\clearpage

\appendix


\begin{table*}[h]
\small
\begin{tabular}{p{1.0\textwidth}}
\toprule
\textbf{Wonder-based Topic Question Generation (w/ NGSS Standard)}\\
\midrule
You are a science teacher (elementary school grade 1). According to the Science Concept, Core Ideas and Learning Outcomes, generate 5 different topics in the form of a short question for elementary grade 1 students.\\
- The generated text should meet the Flesch Kincaid Grade Level for elementary grade 1 students.\\
=== Science Concept ===\\
From Molecules to Organisms: Structures and Processes\\
=== Core Ideas ===\\
Structure and Function • All organisms have external parts. Different animals use their body parts in different ways to see, hear, grasp objects, protect themselves, move from place to place, and seek, find, and take in food, water and air. Plants also have different parts.\\
=== Learning Outcomes ===\\
Students who demonstrate understanding can:
Use materials to design a solution to a human problem by mimicking how plants and/or animals use their external parts to help them survive, grow, and meet their needs.\\
*** System Output ***\\
1. How do animals use their legs to move?\\  
2. What do plants use their leaves for?\\
3. How does a bird use its beak to eat food?\\  
4. Why do turtles have hard shells?\\
5. How do flowers help plants grow?\\
\midrule
\midrule
\textbf{BASE Passage Generation}\\
\midrule
You are a science teacher (elementary school grade 1). Generate a 100-word reading passage around the Wonder Topic for elementary grade 1 students. Mix science and everyday language.\\
=== Wonder Topic ===\\
Why do turtles have hard shells?\\
\midrule
\midrule
\textbf{COGENT-based Content Generation (w/ NGSS Standard)}\\
\midrule
You are a science teacher (elementary school grade 1). Generate a 100-word reading passage around the Wonder Topic to teach students the Science Concept and Core Ideas, to meet the Learning Outcomes. Mix science and everyday language.\\
- The generated text should meet the Flesch Kincaid Grade Level for elementary grade 1 students.\\
=== Wonder Topic ===\\
Why do turtles have hard shells?\\
=== Science Concept ===\\
From Molecules to Organisms: Structures and Processes\\
=== Core Ideas ===\\
What the student needs to learn:\\
Structure and Function • All organisms have external parts. Different animals use their body parts in different ways to see, hear, grasp objects, protect themselves, move from place to place, and seek, find, and take in food, water and air. Plants also have different parts.\\
=== Learning Outcomes ===\\
Students who demonstrate understanding can: 
Use materials to design a solution to a human problem by mimicking how plants and/or animals use their external parts to help them survive, grow, and meet their needs.\\
\bottomrule
\end{tabular}
\caption{Example instructions for wonder question generation, and the BASE and COGENT passage generation. Here the curriculum items are based on the NGSS standard \cite{ngss2013next}. In our experiments, based on human-written passages, we set the word count to be the grade level multiplied by 100. Flesch Kincaid Grade Level \cite{ref_flesch} is used for readability control.}
\label{table-prompt-example}
\vspace{-1.0cm}
\end{table*}

\begin{table*}[h]
\small
\begin{tabular}{p{1.0\textwidth}}
\toprule
\textbf{Curriculum Alignment Scoring}\\
\midrule
I will show you a science reading passage for elementary school students. Rate its curriculum alignment on a 5-point scale (1 = does not align at all, 5 = fully aligned).\\
\textbf{[Curriculum Information]} \\
Grade Level: 3\\
Science Concept: Biological Evolution: Unity and Diversity\\
Core Ideas: Biodiversity and Humans. Populations live in a variety of habitats and change in those habitats affects the organisms living there.\\
Learning Outcomes: Make a claim about the merit of a solution to a problem caused when the environment changes and the types of plants and animals that live there may change.\\
\textbf{[Input Passage Content]}\\
When a forest turns into a field, big changes happen for the plants that live there. Forests are full of tall trees, bushes, and smaller plants that grow in the shade. But fields are open spaces with lots of sunlight and fewer trees. As the forest becomes a field, many of the forest plants can't survive. Trees that once stood tall are cut down or die. The plants that grew in their shade now have too much sun. Some of these plants might dry up or wilt because they're not used to so much light.\\
... ...\\
When habitats change, the living things in them have to change too. Some can't survive, but others find new ways to live. Nature is always changing, and plants and animals are always trying to keep up.\\
\textbf{[System Output]}\\
Alignment Score: 5\\
\midrule
\midrule
\textbf{Curriculum Item Categorization}\\
\midrule
Classify the science reading passage for elementary school students into one of the following types according to the curriculum definition. Give me the type label.\\
\textbf{[Curriculum Item Categories]}\\
"Type": "A",\\
"Concept": "From Molecules to Organisms: Structures and Processes",\\
"Core Ideas": "Structure and Function • All organisms have external parts. Different animals use their body parts in different ways to see, hear, grasp objects, protect themselves, move from place to place, and seek, find, and take in food, water, and air. Plants also have different parts",\\
"Learning Outcomes": "Use materials to design a solution to a human problem by mimicking how plants and/or animals use their external parts to help them survive, grow, and meet their needs.",\\
"Type": "B",\\
"Concept": "From Molecules to Organisms: Structures and Processes",\\
"Core Ideas": "Growth and Development of Organisms • Adult plants and animals can have young. In many kinds of animals, parents and the offspring themselves engage in behaviors that help the offspring to survive.",\\
"Learning Outcomes": "Read texts and use media to determine patterns in behavior of parents and offspring that help offspring survive.",\\
...\\
...\\
"Type": "G",\\
"Concept": "From Molecules to Organisms: Structures and Processes",\\
"Core Ideas": "Organization for Matter and Energy Flow in Organisms • Plants acquire their material for growth chiefly from air and water.",\\
"Learning Outcomes": "Support an argument that plants get the materials they need for growth chiefly from air and water.",\\
\textbf{[Input Passage Content]}\\
Cats have special hairs called whiskers. These whiskers are not like normal fur. They are thick and stiff. Whiskers grow on a cat's face and legs. They help cats in many ways. Cats use whiskers to feel things around them. This helps them move in the dark. Whiskers can sense air movement too. This tells cats if something is nearby. When hunting, whiskers help cats know if they can fit through small spaces. Cats also use whiskers to show how they feel. If a cat is happy, its whiskers point forward. When scared, the whiskers go back. Whiskers are very important for cats. They help cats stay safe and find food.\\
\textbf{[System Output]}\\
Predicted Type: A\\
\bottomrule
\end{tabular}
\caption{Example instructions for curriculum alignment scoring and curriculum item categorization.}
\label{table-inst-alignment}
\end{table*}

\begin{table*}[h]
\small
\begin{tabular}{p{1.0\textwidth}}
\toprule
\textbf{Comprehensibility Assessment}\\
\midrule
I will show you a science reading passage for elementary school students.
Rate its comprehensibility on readability, correctness, coherence, and engagement. Give me the score of each aspect in a 5-point Likert.\\
\textbf{[Target Grade Level]}\\
Grade 3\\
\textbf{[Comprehensibility Aspects]}\\
- Readability: How easily the passage can be read and understood\\
- Correctness: The accuracy of factual content about the topic\\
- Coherence: The consistency and unity between the content and the topic\\
- Engagement: To what extent the ``wonder-based'' topic and passage capture and maintain readers' interest\\
\textbf{[Input Passage Content]}\\
\textbf{Topic}: What happens to plants if a forest becomes a field?\\
When a forest turns into a field, big changes happen for the plants that live there. Forests are full of tall trees, bushes, and smaller plants that grow in the shade. But fields are open spaces with lots of sunlight and fewer trees. As the forest becomes a field, many of the forest plants can't survive. Trees that once stood tall are cut down or die. The plants that grew in their shade now have too much sun. Some of these plants might dry up or wilt because they're not used to so much light.\\
... ...\\
When habitats change, the living things in them have to change too. Some can't survive, but others find new ways to live. Nature is always changing, and plants and animals are always trying to keep up.\\
\textbf{[System Output]}\\
Readability: 5, Correctness: 5, Coherence: 5, Engagement: 5\\
\bottomrule
\end{tabular}
\caption{Example instructions for the comprehensibility assessment.}
\label{table-inst-comprehensibility}
\vspace{-1.0cm}
\end{table*}

\begin{table*}[h]
\small
\begin{tabular}{p{1.0\textwidth}}
\toprule
\textbf{Text Readability Assessment}\\
\midrule
\textbf{Topic:} What do plants use roots for?\\
\textbf{Grade Level:} 2\\
Plants have roots that grow under the ground. Roots are very important for plants. They help plants get water from the soil. The roots also hold the plant in place. This keeps the plant from falling over. Some roots can even store food for the plant. Roots come in different shapes and sizes. Some are long and thin, while others are short and thick. Plants use their roots to drink water, just like we use our mouths. The roots also help plants get nutrients from the soil. These nutrients help the plant grow big and strong. Without roots, plants would not be able to live and grow.\\
\textbf{Readability Scores:}\\
Flesch Reading Ease: 96.28,
Flesch Kincaid Grade Level: 2.0,\\
Gunning Fog Index: 3.93,
Automated Readability Index: 4.1,
Coleman Liau Index: 6.06\\
\midrule
\midrule
\textbf{Topic:} What does a fish's gills do?\\
\textbf{Grade Level:} 4\\
Have you ever wondered how fish can breathe underwater? Unlike us, fish don't have lungs to take in air. Instead, they have special organs called gills that help them get oxygen from the water around them. These gills are a great example of how animals have special body parts that help them survive in their environment. Fish gills are like tiny, feathery structures on the sides of a fish's head. They're hidden under flaps called gill covers, which protect them. When a fish opens its mouth, water flows in and passes over the gills. As this happens, the gills grab oxygen from the water, just like our lungs grab oxygen from the air we breathe. The gills are made up of many thin, flat sheets called gill filaments. These filaments are packed with tiny blood vessels. As water moves over the filaments, oxygen from the water passes into the blood vessels. At the same time, carbon dioxide, which is a waste gas, moves from the blood into the water. This swap of gases is how fish \"breathe\" underwater. Fish gills do more than just help with breathing. They also play a part in getting rid of waste from the fish's body and keeping the right balance of salt and water. Some fish that live in salt water have special cells in their gills that can remove extra salt from their bodies. Different types of fish have gills that are slightly different, depending on where they live and how they behave. For example, fast-swimming fish like tuna have gills that can take in more oxygen quickly. This helps them keep up their speedy swimming. On the other hand, fish that live in waters with less oxygen, like some catfish, have gills that are really good at grabbing what little oxygen is there. It's amazing how fish gills work so well for life underwater. They're a perfect fit for the fish's watery home, just like our lungs are a perfect fit for life on land. This shows how animals have body parts that are just right for where they live and what they need to do to survive.\\
\textbf{Readability Scores:}\\
Flesch Reading Ease: 81.12,
Flesch Kincaid Grade Level: 5.8,\\
Gunning Fog Index: 7.44,
Automated Readability Index: 7.7,
Coleman Liau Index: 8.0\\
\bottomrule
\end{tabular}
\caption{Two evaluation examples of the text readability assessment.}
\label{table-inst-readability}
\end{table*}

\begin{table*}[h]
\small
\begin{tabular}{p{1.0\textwidth}}
\toprule
\textbf{[Curriculum Information]} \\
\textbf{Grade level}: 3\\
\textbf{Science concept}: Biological Evolution: Unity and Diversity\\
\textbf{Core ideas}: Biodiversity and Humans. Populations live in a variety of habitats and change in those habitats affects the organisms living there.\\
\textbf{Learning objectives}: Make a claim about the merit of a solution to a problem caused when the environment changes and the types of plants and animals that live there may change.\\
\textbf{Topic Question}: Why do birds migrate?\\
\midrule
\midrule
\textbf{[Human-written Passage]} \\
\textcolor{blue}{Migration is very hard work, but the work has big benefits, such as food—lots of it!} Many birds that migrate north in spring arrive just as huge numbers of insects are hatching. It’s a buggy banquet full of protein for both hungry arriving adults and the young birds soon to be hatched. Growing chicks especially need lots of insect protein to grow big and strong. Spring also brings new seeds, fruits, and nectar-filled flowers. Thanks to all this food, many migratory birds raise more young than birds that stay in warm places all year long. \textcolor{blue}{Migration is also dangerous. Storms can kill birds or send them off course.} Many migrating birds crash into windows and brightly lit tall buildings. One of the biggest problems for any migrating bird is the loss of habitat. \textcolor{blue}{Birds need healthy habitats everywhere they spend time throughout the year.} That includes the places they nest and raise their young, the places they spend the winter, and the places they stop along their migratory journeys to rest and feed. \textcolor{blue}{They need wild unbuilt places like woods, beaches, grasslands, and wetlands.} When natural areas are replaced by roads, homes, shopping centers, farms, and other human-made structures, there is less of the healthy habitat birds need throughout the year. \textcolor{blue}{A healthy habitat is also a place where there are plenty of native plants—the kinds of plants that grow naturally in a particular area.} Those plants provide more nutritious fruit and host many more insects than plants that are from other places. When people replace native plants with nonnative plants, the amount of the right kind of food available to birds is reduced. \textcolor{blue}{Many people want to help migratory birds.} In all kinds of communities, people are planting native plants that provide food and places for hummingbirds and songbirds to nest and rest. They are working to protect grasslands for the hawks that hunt there and other birds that nest there. Beaches, wetlands, and bodies of water are being cleaned and protected to provide healthy habitats for shorebirds, wading birds, and swimming birds like ducks and geese. \\
\midrule
\midrule
\textbf{[BASE-generated Passage]} \\
Have you ever seen a group of birds flying high in the sky, all moving in the same direction? They might be migrating! \textcolor{blue}{Migration is when birds travel from one place to another at certain times of the year.} But why do they do this? \textcolor{blue}{One big reason birds migrate is to find food.} When winter comes, many insects hide away and plants stop growing in colder places. Just like how we go to the grocery store when our refrigerator is empty, birds fly to warmer places where they can find plenty of tasty insects, seeds, and berries to eat. Some birds, like hummingbirds, travel thousands of miles just to find their favorite flower nectar! \textcolor{blue}{Another reason birds migrate is to stay safe and comfortable.} Cold weather can be dangerous for birds – their tiny bodies need lots of energy to stay warm. By flying to warmer places, they can save their energy for important things like finding food and building nests. \textcolor{blue}{Many birds also migrate to find better places to raise their babies, where there are fewer predators and more materials to build cozy nests.} When the weather changes again, many birds fly back to where they started. This cycle happens every year. Some clever birds, like geese, even use the sun and stars to help them find their way, just like a natural compass! So, the next time you see birds flying in a group, think about their amazing journey. They might be heading to their warm winter vacation or coming back home for spring! \\
\midrule
\midrule
\textbf{[COGENT-generated Passage]} \\
Have you ever wondered why some birds disappear in the fall and come back in the spring? \textcolor{blue}{This is called migration. Birds fly long distances to find better places to live and eat. It's like taking a vacation, but for birds, it's about survival.} \textcolor{blue}{Many birds migrate because the weather changes.} When it gets cold, some birds can't find enough food. Insects hide, and plants stop growing. So, these birds fly to warmer places where they can find more to eat. When spring comes, they fly back to their old homes. \textcolor{blue}{Birds face many challenges during migration.} Flying long distances is hard work. \textcolor{blue}{They need to eat a lot before they leave to have enough energy for the trip.} Some birds fly over big oceans or deserts where they can't stop to rest or eat. They have to be strong and healthy to make it. \textcolor{blue}{Another problem is that the places birds visit are changing.} People build houses and cities where birds used to stop and rest. This makes it harder for birds to find safe places to sleep and eat during their journey. Some birds get confused by bright lights in cities and fly into buildings. \textcolor{blue}{Weather can also be a big problem for migrating birds.} Strong winds and storms can blow them off course. They might end up in places they don't know, where it's hard to find food. Climate change is making some areas too hot or dry for birds, so they have to find new places to go. \textcolor{blue}{Birds have amazing ways to deal with these challenges.} They can fly very high to catch good winds that help them travel faster. Many birds migrate in big groups, which helps them stay safe from predators. Some birds can even sleep while they're flying! \textcolor{blue}{People are trying to help migrating birds.} They're making special areas called wildlife refuges where birds can rest and eat safely. Some cities turn off bright lights at night during migration season to help birds fly safely. Scientists are studying birds to learn more about how to protect them. \textcolor{blue}{You can help migrating birds too.} Planting trees and flowers in your yard gives birds places to rest and eat. Keeping cats inside helps protect birds from getting hurt. By taking care of the places where birds live and stop during migration, we can make sure they have safe journeys every year. \\
\bottomrule
\end{tabular}
\caption{Three passages upon the same curriculum information. Text spans in blue highlight where scientific concepts and ideas are introduced. It also marks explanations of scientific phenomena that directly connect to learning objectives.}
\end{table*}

\end{document}